\documentclass{article}
\usepackage{graphicx} 
\usepackage{wasysym}
\usepackage{subcaption}

\usepackage{microtype}
\usepackage{booktabs} 

\usepackage{hyperref}
\usepackage{amsthm}
\usepackage{amssymb}
\usepackage{mathtools}
\usepackage{changepage}
\usepackage{tabularx}
\usepackage{float}

\AtBeginDocument{%
  }

\title{ Improving Predictions on Highly Unbalanced Data\\ Using Open Source Synthetic Data Upsampling}
\author{
  Ivona Krchova \\
  MOSTLY AI \\
  \texttt{ivona.krchova@mostly.ai}
    \and
  Michael Platzer \\
  MOSTLY AI \\
  \texttt{michael.platzer@mostly.ai}
  \and
  Paul Tiwald \\
  MOSTLY AI \\
  \texttt{paul.tiwald@mostly.ai}
}

\date{}  %

\begin{document}
\maketitle

\begin{abstract}
Unbalanced tabular data sets present significant challenges for predictive modeling and data analysis across a wide range of applications. In many real-world scenarios, such as fraud detection, medical diagnosis, and rare event prediction, minority classes are vastly underrepresented, making it difficult for traditional machine learning algorithms to achieve high accuracy. These algorithms tend to favor the majority class, leading to biased models that struggle to accurately represent minority classes. Synthetic data holds promise for addressing the under-representation of minority classes by providing new, diverse, and highly realistic samples. This paper presents a benchmark study on the use of AI-generated synthetic data for upsampling highly unbalanced tabular data sets.

We evaluate the effectiveness of an open-source solution, the Synthetic Data SDK by MOSTLY AI, which provides a flexible and user-friendly approach to synthetic upsampling for mixed-type data. We compare predictive models trained on data sets upsampled with synthetic records to those using standard methods, such as naive oversampling and SMOTE-NC. Our results demonstrate that synthetic data can improve predictive accuracy for minority groups by generating diverse data points that fill gaps in sparse regions of the feature space. We show that upsampled synthetic training data consistently results in top-performing predictive models, particularly for mixed-type data sets containing very few minority samples.

\end{abstract}

\section{Introduction}

AI-generated synthetic data, which we refer to as synthetic data in the following, is created by training a generative model on the original data set. In the inference phase, the generative model creates statistically representative synthetic records from scratch. The use of synthetic data has gained increasing importance in various privacy-sensitive industries due to its primary use case of enhancing data privacy. Beyond privacy, synthetic data offers the possibility to modify and tailor data sets to specific needs. In this research paper, we investigate the potential of synthetic data to improve the performance of machine learning algorithms on data sets with unbalanced class distributions, specifically through the synthetic upsampling of minority classes. \\
Class imbalance is a common problem in many real-world tabular data sets, where the number of samples in one or more classes is significantly lower than in others. Such imbalances can lead to poor prediction performance for the minority classes, which are often of greatest interest in applications such as fraud detection or extreme insurance claims. Traditional upsampling methods, such as naive oversampling or SMOTE \cite{Chawla_2002}, have shown some success in mitigating this issue. Despite their widespread use, these methods often have limitations and can introduce biases that negatively impact model performance. Poor model performance was highlighted in a study that demonstrated classical upsampling techniques, including SMOTE, are ineffective when applied in combination with traditional machine learning algorithms \cite{creditscoringstudy}.
\\ 
Naive oversampling mitigates class imbalance effects by simply duplicating minority class examples. Due to this strategy, it bears the risk of overfitting the model to the training data, resulting in poor generalization in the inference phase. SMOTE, on the other hand, generates new records by interpolating between existing minority class samples, leading to higher diversity. However, SMOTE’s ability to increase diversity is also limited when the absolute number of minority records is very low. This is especially true when generating samples for mixed-type data sets containing categorical columns. For mixed-type data sets, SMOTE-NC \cite{imblearn_smotenc} is commonly used as an extension for handling categorical columns. SMOTE-NC may not work well with non-linear decision boundaries, as it only linearly interpolates between minority records. This can lead to SMOTE-NC examples being generated in an “unfavorable” region of feature space, far from where additional samples would help the predictive model to place a decision boundary.\\
In response, synthetic data upsampling with generative AI models has emerged as a practical alternative \cite{overlapclass}.
The strength of upsampling minority classes with AI-generated synthetic data is that the generative model is not limited to upsampling or interpolating between existing minority samples. Most AI-based generators can create realistic synthetic data examples in any region of feature space and thus considerably increase diversity. Because they are not tied to existing minority samples, AI-based generators can also leverage and learn from properties of parts of the majority class that are transferable to minority examples.

An additional strength of using AI-based upsampling is that it can be easily extended to more complex data structures, such as sequential data, where not just one but many rows in a data set belong to a single data subject. This aspect of synthetic data upsampling, however, is out of the scope of this study.

In our study, we used the Synthetic Data SDK by MOSTLY AI \cite{mostlyai}, an open-source Python library, because of its ease of use and efficient local operation compared to LLM-based approaches \cite{LLMoversampling}. This library provides high-quality upsampling capabilities without significant computational overhead, making it a practical choice for real-world applications such as bank customer scoring \cite{scoring}.

\section{Experimental Setup}

\begin{figure}[H]
    \centering
    \includegraphics[width=\columnwidth]{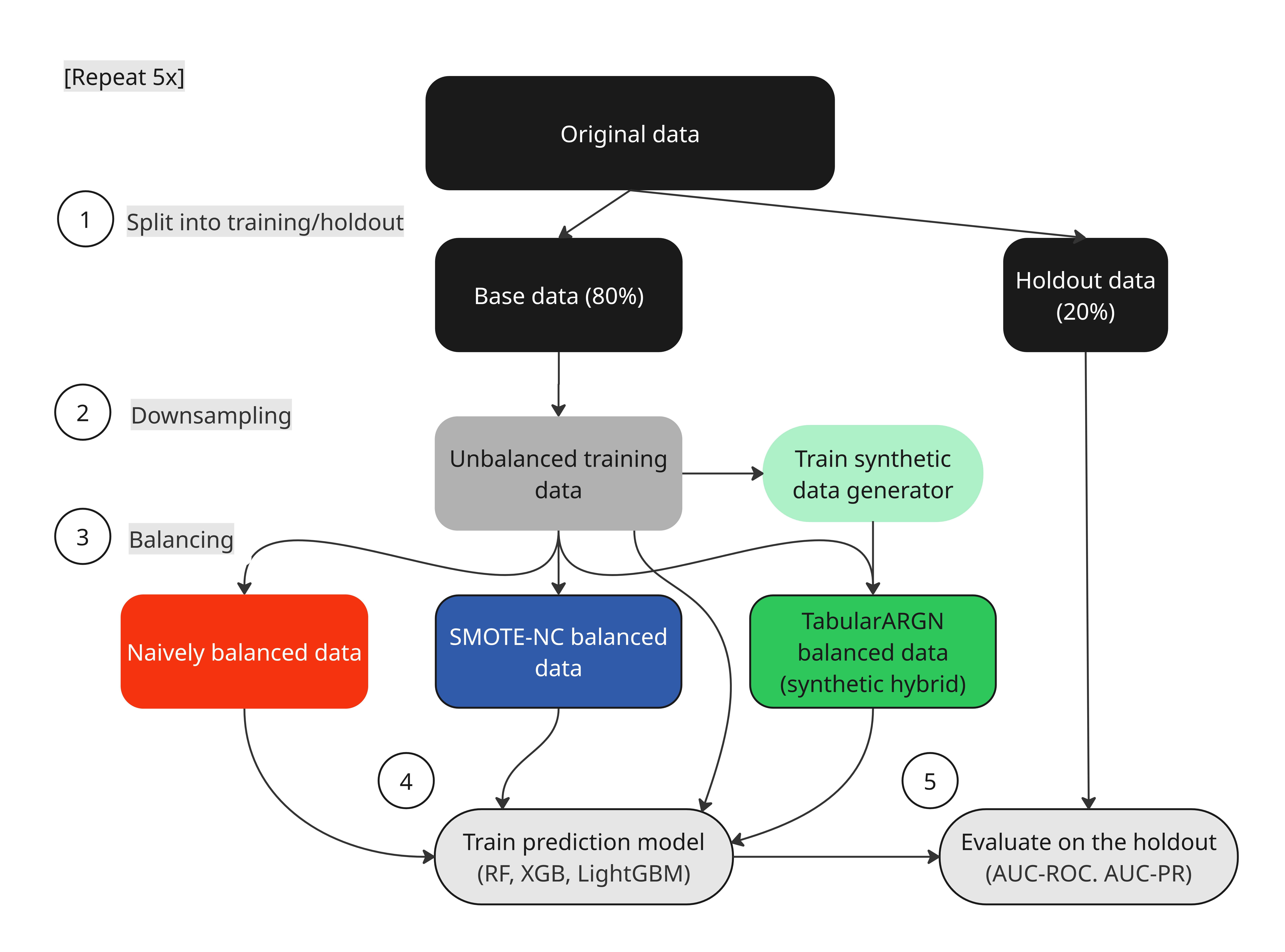}
    \caption{Experimental Setup: (1) We split the original data set into a base data set and a holdout. (2) Strong imbalances are introduced in the base data set by downsampling the minority classes. (3) We test different mitigation strategies: balancing through naive upsampling, SMOTE-NC upsampling, and upsampling with TabularARGN (the synthetic hybrid data set). (4) We train LGBM, RandomForest, and XGB classifiers on the balanced and unbalanced training data. (5) We evaluate the properties of the upsampling techniques by measuring the performance of the trained classifier on the holdout set. Steps 1-5 are repeated 5 times and we report the mean AUC-ROC as well as the AUC-PR.}
    ~\label{fig:exp-setup}
\end{figure}

For each data set used in our experiments, we run through the following steps (see fig.\ \ref{fig:exp-setup})

\begin{enumerate}

\item We split each original data set into a base set and a holdout set using a five-fold stratified sampling approach to ensure that each class is represented proportionally.

\item All original data sets have a binary target column and only a moderate imbalance, with the fraction of the minority class ranging from 12\% to 24\% (see Table~\ref{tab:datasetproperties} for details). We artificially induce different levels of strong imbalance in the base set by randomly down-sampling the minority class, resulting in unbalanced training data sets with minority fractions of 0.05\%, 0.1\%, 0.2\%, 0.5\%, 1\%, 2\%, and 5\%.

\item To mitigate the strong imbalance in the training data sets, we apply three different upsampling techniques:
\begin{itemize}
\item Naive oversampling (red box in Fig.\ref{fig:exp-setup}): duplicating existing examples of the minority class\cite{imblearn_naive, scikit-learn}.
\item SMOTE-NC (blue box in Fig.\ref{fig:exp-setup}): applying the SMOTE-NC upsampling technique\cite{imblearn_smotenc, scikit-learn}.
\item TabularARGN (green box in Fig.\ref{fig:exp-setup}): enriching unbalanced training data with AI-generated synthetic data. It consists of the original training data (including majority samples and a limited number of minority samples) plus additional synthetic minority samples created using an AI-based synthetic data generator. This generator is trained on the highly unbalanced training data set. In this study, we utilize the Synthetic Data SDK by MOSTLY AI\cite{mostlyai}, which includes built-in support for rebalancing imbalanced data sets and is based on the TabularARGN framework~\cite{tabularargn}. Detailed instructions on how to use upsampling within the SDK can be found in the online tutorial\footnote{\url{https://github.com/mostly-ai/mostlyai/tree/main/docs/tutorials/rebalancing}}.
\end{itemize}

In all cases, we upsample the minority class to achieve a 50:50 balance between the majority and minority classes, resulting in the naively balanced, SMOTE-NC balanced, and balanced hybrid data sets.

\item We assess the benefits of the different upsampling techniques by training three popular classifiers—Random Forest, XGBoost, and LightGBM—on the balanced data sets. Additionally, we train the classifiers on the heavily unbalanced training data sets to serve as a baseline for evaluating predictive model performance.

\item The classifiers are evaluated on the holdout set, and we calculate both the AUC-ROC and AUC-PR scores across all upsampling techniques and initial imbalance ratios for all five folds. We report the average scores over five different samplings and model predictions. We choose AUC metrics to eliminate dependencies on thresholds, as seen in metrics such as the F1 score.

\end{enumerate}

\section{Data Sets and Results}
We use three publicly available data sets of varying sizes (Table~\ref{tab:datasetproperties}): \textit{Adult}~\cite{adult}, \textit{Credit Card}~\cite{credit}, and \textit{Insurance}~\cite{insurance}. All data sets are of mixed type (categorical and numerical features) and have a binary, i.e., categorical target column.

In step 2 (Fig.~\ref{fig:exp-setup}), we downsample the minority classes to induce strong imbalances. For the smaller data sets with approximately 30{,}000 records, downsampling to minority-class fractions of 0.1\% results in extremely low numbers of minority records. The downsampled Adult and Credit Card unbalanced training data sets contain as few as 19 and 18 minority records, respectively. This scenario mimics situations where data is limited and extreme cases occur rarely. Such setups create significant challenges for predictive models, as they may struggle to make accurate predictions and generalize well to unseen data.

Please note that the holdout sets, on which the trained predictive models are evaluated, are not subject to extreme imbalances, as they are sampled from the original data before downsampling is applied. The imbalance ratios of the holdout sets are moderate and range from 12\% to 24\%.

In the evaluation, we report both AUC-ROC and AUC-PR, due to the moderate but inhomogeneous distribution of minority fractions in the holdout sets. While AUC-ROC is a very popular and expressive metric, it is known to be overly optimistic for unbalanced classification problems~\cite{pr}. The AUC-ROC considers both classes, making it susceptible to neglecting the minority class, whereas the AUC-PR focuses on the minority class, as it is based on precision and recall.

\begin{table}[H]
  \centering
\small
  \begin{tabularx}{\columnwidth}{l r r r  r r r}
    {\small\textit{Dataset}}
    & {\small \textit{Minority}}
     & {\small \textit{N}}
     & {\small \textit{C}}
     & {\small \textit{Target}}
     & {\small \textit{Original}}
    & {\small \textit{\# minority}} \\
    \midrule
    Adult & 24 \% & 5 & 8 & income & 32 561  & 19 \\
    Credit card & 22 \% & 21 & 3 & default payment & 30 000  & 18 \\
    Insurance & 12 \% & 6 & 4 & Response & 381 109  & 268 \\
  \end{tabularx}
  \caption{Data sets properties used in this study. \textit{Minority} denotes the fraction of the minority class in the original data set. \textit{N} and \textit{C} is the number of numeric and categorical features, respectively. \textit{Target} is the target column the classifiers are trained to predict. \textit{Original} is the number of records in the original data set. \textit{\# minority} denotes the number of minority records in the unbalanced training data with a 0.1\% fraction of the minority class.}~\label{tab:datasetproperties}
\end{table}


\subsection{Adult} \label{sec:adult}
\begin{figure}[htb]
    \centering
    \begin{subfigure}{0.49\textwidth}
        \includegraphics[width=\linewidth]{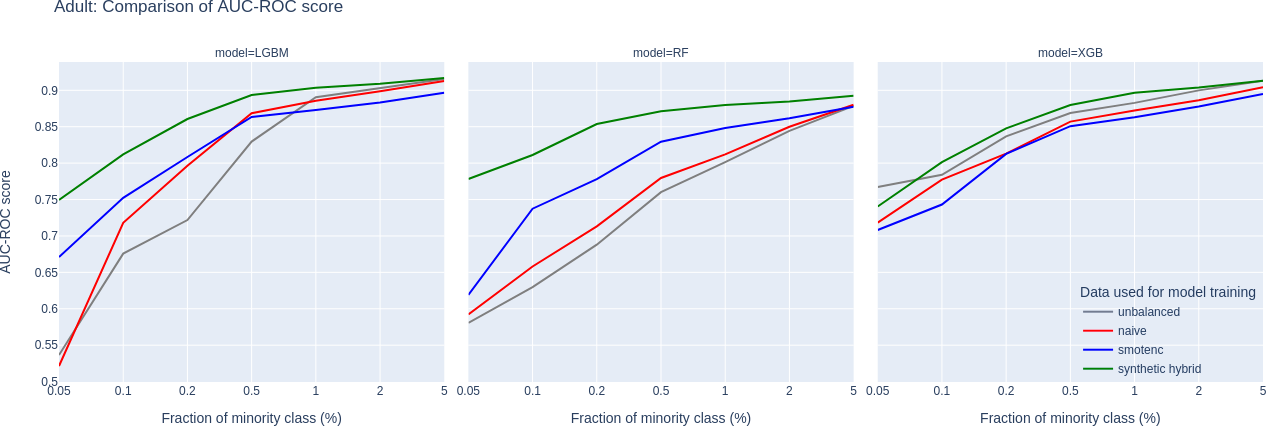}
        \caption{AUC-ROC}
    \end{subfigure}
    \hfill
    \begin{subfigure}{0.49\textwidth}
        \includegraphics[width=\linewidth]{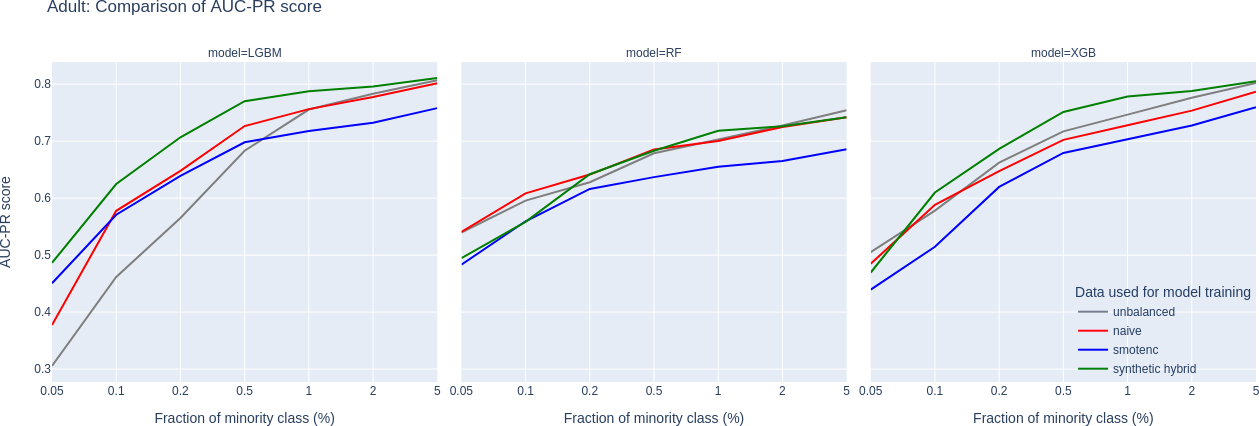}
        \caption{AUC-PR}
    \end{subfigure}
    \caption{AUC-ROC (a) and AUC-PR (b) of classifiers LGBM, RandomForest (RF), and XGB trained on the \textit{Adult} data set to predict the target feature \textit{income}. The classifiers are trained on unbalanced data sets (grey) and data sets that are upsampled naively (red), with the SMOTE-NC algorithm (blue), and with TabularARGN synthetic records (green). AUC values are reported for different fractions of the minority class in the unbalanced training data (x-axis).}
    \label{fig:adult-auc-roc-pr}
\end{figure}


The largest differences between upsampling techniques are observed in the AUC-ROC when balancing training sets with a substantial class imbalance of 0.05\% to 0.5\%. This scenario involves a very limited number of minority samples, with as few as 19 for the Adult unbalanced training data set.

For the RF and LGBM classifiers trained on the balanced hybrid data set, the AUC-ROC is higher than those obtained with other upsampling techniques. Differences can reach up to 0.2 (RF classifier, minority fraction of 0.05\%) between the TabularARGN-based upsampling and the second-best method.

The AUC-PR shows similar, yet less pronounced, differences. LGBM and XGB classifiers trained on the balanced hybrid data set perform best across almost all minority fractions. Interestingly, results for the RF classifier are mixed: upsampling with synthetic data does not always lead to better performance, but it is always among the best-performing methods.

While synthetic data upsampling improves results for most minority fractions with the XGB classifier as well, the differences in performance are less pronounced. Notably, the XGB classifier trained on the highly unbalanced training data performs surprisingly well. This suggests that the XGB classifier is better suited for handling unbalanced data.

\begin{figure}[H]
    \centering
    \begin{subfigure}{0.49\textwidth}
        \includegraphics[width=\linewidth]{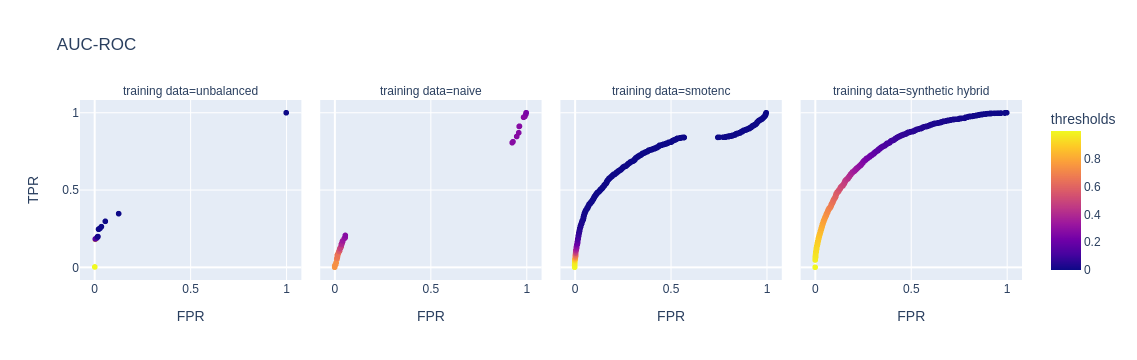}
        \caption{ROC Curve}
    \end{subfigure}
    \hfill
    \begin{subfigure}{0.49\textwidth}
        \includegraphics[width=\linewidth]{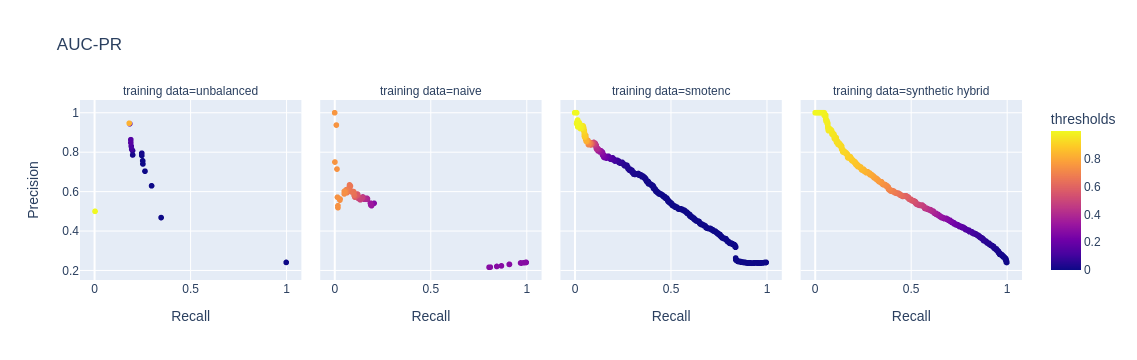}
        \caption{PR Curve}
    \end{subfigure}
    \caption{ROC (a) and PR (b) curve of the Light GBM classifier trained on different versions of the adult data set (from left to right): unbalanced training data (minority class fraction 0.1\%), naively upsampled training data, training data upsampled with SMOTE-NC, and training data enriched with TabularARGN synthetic records (synthetic hybrid). Every point on the plots corresponds to a specific prediction threshold of the classifier.}
    \label{fig:adult-roc-pr-curve}
\end{figure}

The reason for the performance differences in the AUC-ROC and AUC-PR is the low diversity and resulting overfitting when using naive or SMOTE-NC upsampling. These effects are visible, for example, in the ROC and PR curves of the LGBM classifier for a minority fraction of 0.1\% (Fig.~\ref{fig:adult-roc-pr-curve}).

Each point on these curves corresponds to a specific prediction threshold of the classifier. The set of threshold values is determined by the variance of probabilities predicted by the models when evaluated on the holdout set. For both the highly unbalanced training data and the naively upsampled data, we observe very low diversity, with more than 80

In the PR curve plot, this leads to an accumulation of points in the area with high precision and low recall, meaning that the model is very conservative in making positive predictions and only predicts a positive class when it is very confident the data point belongs to the minority class. This demonstrates the effect of overfitting on a few samples in the minority group.

SMOTE-NC results in a much smoother PR curve, which, however, still contains discontinuities and has a large segment where precision and recall change rapidly with small changes in the prediction threshold.

The TabularARGN hybrid data set offers high diversity during model training, resulting in almost every holdout sample being assigned a unique probability of belonging to the minority class. Both the ROC and PR curves are smooth and have a threshold of approximately 0.5 at the center, which is closest to the perfect classifier.

\begin{figure}[H]
    \centering
    \includegraphics[width=\columnwidth]{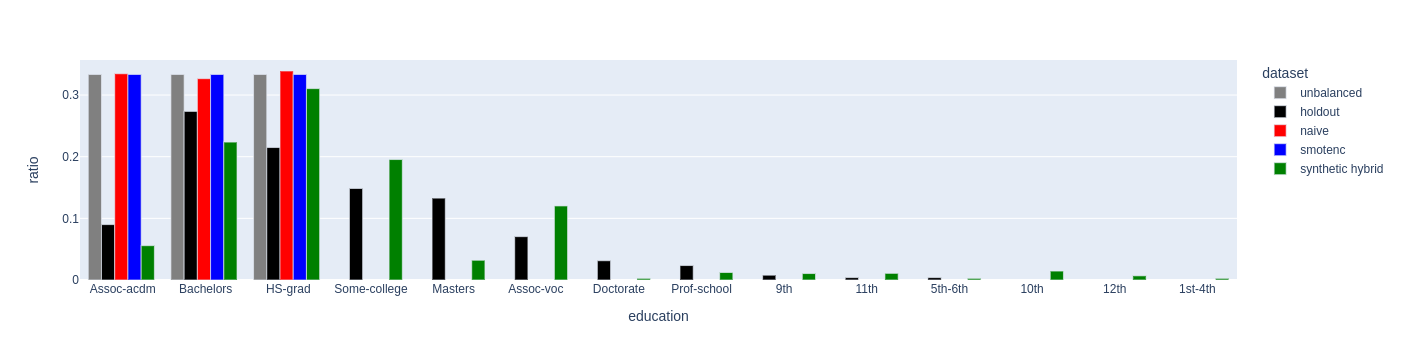}
    \caption{Distribution of the feature \textit{education} for the female subgroup (\textit{sex} equals \textit{female}) of the minority class (\textit{income} equals \textit{high}). The distributions of the unbalanced (grey - minority class fraction of 0.1\%), the naively upsampled (red), and the SMOTE-NC upsampled (blue) considerably differ from the holdout distribution. Only the data set upsampled with TabularARGN synthetic records (synthetic hybrid) recovers the holdout distribution to a satisfactory degree and captures its diversity.}
    \label{fig:adult-females}
\end{figure}

The limited power in creating diverse samples when the minority class is severely underrepresented stems from naive upsampling and SMOTE-NC being restricted to duplicating and interpolating between existing minority samples. Both methods are confined to a limited region in feature space. Upsampling with TabularARGN-based synthetic minority samples, on the other hand, can, in principle, populate any region in feature space and can leverage properties of the majority samples that are transferable to minority examples, resulting in more diverse and realistic synthetic minority samples.

We analyze the difference in diversity by further “drilling down” into the minority class (feature \textit{income} equals \textit{high}) and comparing the distribution of the feature \textit{education} for the female subgroup (feature \textit{sex} equals \textit{female}) in the upsampled data sets (Fig.~\ref{fig:adult-females}). For a minority fraction of 0.1\%, this results in only three female minority records. Naive upsampling and SMOTE-NC have great difficulty generating diversity in such settings. Both just duplicate the existing categories—\textit{Bachelors}, \textit{HS-grade}, and \textit{Assoc-acdm}—resulting in a strong distortion of the distribution of the \textit{education} feature compared to the holdout data set. The distribution of the TabularARGN synthetic hybrid data set also has some imperfections, but it recovers the holdout distribution to a much greater degree. Many more \textit{education} categories are populated and, with a few exceptions, the frequencies from the holdout data set are recovered to a satisfactory level. This ultimately leads to greater diversity in the hybrid data set than in either the naively balanced or SMOTE-NC balanced ones.

We quantitatively assess diversity using the Shannon entropy \cite{shannon}, which measures variability within a data set, particularly for categorical data. It provides an indication of how uniformly the different categories of a specific feature are distributed within the data set.

The Shannon Entropy (SE) of a specific feature is defined as
\begin{equation}
SE = - \sum_{i}(p(i) \cdot \log_{2}(p(i))),
\end{equation}
where $p(i)$ represents the probability of occurrence, i.e.\ the relative frequency of category $i$. $SE$ ranges from $0$ to $log_2(N)$, where $N$ is the total number of categories. A value of $0$ indicates maximum certainty with only one category, while higher entropy implies greater diversity and uncertainty, indicating comparable probabilities $p(i)$ across categories.

In Table~\ref{tab:entropy}, we report the Shannon entropy for different features and subgroups of the \textit{high-income} population. In all cases, data diversity is highest in the holdout data set. The downsampled training data set (\textit{unbalanced}) has a greatly reduced SE, especially when focusing on the small group of \textit{high-income women}. Naive and SMOTE-NC upsampling cannot recover any of the diversity present in the holdout, as both are limited to the categories found in the minority class. In line with the results presented above, synthetic data recovers the SE, that is, the diversity of the holdout data set, to a large degree.

\begin{table}
  \centering
  \small
  \begin{tabular}{c c r r r r r }
    {\small\textit{Subgroup}}
    & {\small\textit{feature(s)}}
    & {\small \textit{holdout}}
     & {\small \textit{unbalanced}}
     & {\small \textit{naive}}
     & {\small \textit{smotenc}}
     & {\small \textit{synth. hybrid}}\\
    \midrule
    {\small \female} & {\small education} & {\small 2.77}	& {\small 1.58}	& {\small 1.58} & {\small 1.58} & {\small 2.65} \\
    {\small \female} & {\small all (avg.)} & {\small 1.45} & {\small 0.65}	& {\small 0.65}  & {\small 0.65} & {\small 1.28} \\
    {\small \female \space and \male} & {\small all (avg.)}  & {\small 1.44}	& {\small 1.34}	& {\small 1.34} & {\small 0.92} & {\small 1.38} \\
  \end{tabular}
  \caption{Shannon entropy for different features and subgroups of the \textit{high-income} population of the \textit{Adult} data set. Feature \textit{all (avg.)} refers to the average across the SEs of all individual features. Numeric features are binned into deciles before calculating the SE.}~\label{tab:entropy}
\end{table}

\subsection{Credit Card}

The Credit Card data set has similar properties to the Adult data set. The number of records, features, and the original moderate imbalance are comparable. This again results in a very small number of minority records (18) after downsampling to a 0.1\% minority fraction. The main difference between them is that the Credit Card data set consists of more numeric features.

\begin{figure}[H]
    \centering
    \begin{subfigure}{0.49\textwidth}
        \includegraphics[width=\linewidth]{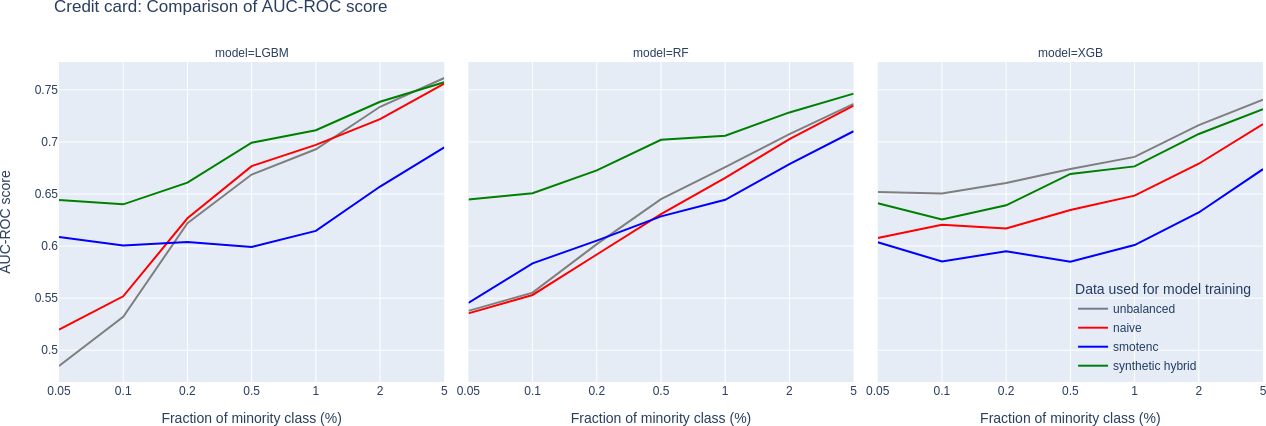}
        \caption{AUC-ROC}
    \end{subfigure}
    \hfill
    \begin{subfigure}{0.49\textwidth}
        \includegraphics[width=\linewidth]{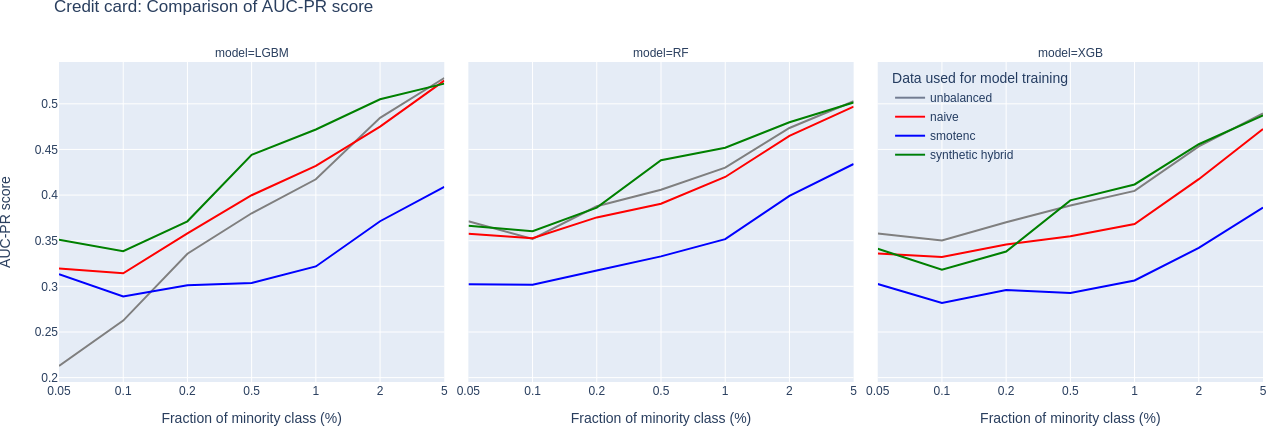}
        \caption{AUC-PR}
    \end{subfigure}
    \caption{AUC-ROC (a) and AUC-PR (b) of classifiers LGBM, RandomForest (RF), and XGB trained on the \textit{Credit Card} data set to predict the target feature \textit{default payment}. The classifiers are trained on unbalanced data sets (grey) and data sets that are upsampled naively (red), with the SMOTE-NC algorithm (blue), and with TabularARGN synthetic records (green). AUC values are reported for different fractions of the minority class in the unbalanced training data (x-axis).}
    \label{fig:credit-auc-roc-pr}
\end{figure}

The performance of different upsampling techniques on the unbalanced Credit Card training data set shows results similar to the Adult data set (Fig.~\ref{fig:credit-auc-roc-pr}). AUC-ROC and AUC-PR for both LGBM and RF classifiers improve over naive upsampling and SMOTE-NC when using the TabularARGN hybrid data set. Again, the performance of the XGB model is more comparable across the different balanced data sets, and we find very good performance even for the highly unbalanced training data set. Here, too, the synthetic hybrid data set is always among the best-performing upsampling techniques.

Interestingly, SMOTE-NC performs worst across almost all metrics. This is surprising, because we would expect this data set, consisting mainly of numerical features, to be favorable for the SMOTE-NC upsampling technique.

\subsection{Insurance}

\begin{figure}[H]
    \centering
    \begin{subfigure}{0.49\textwidth}
        \includegraphics[width=\linewidth]{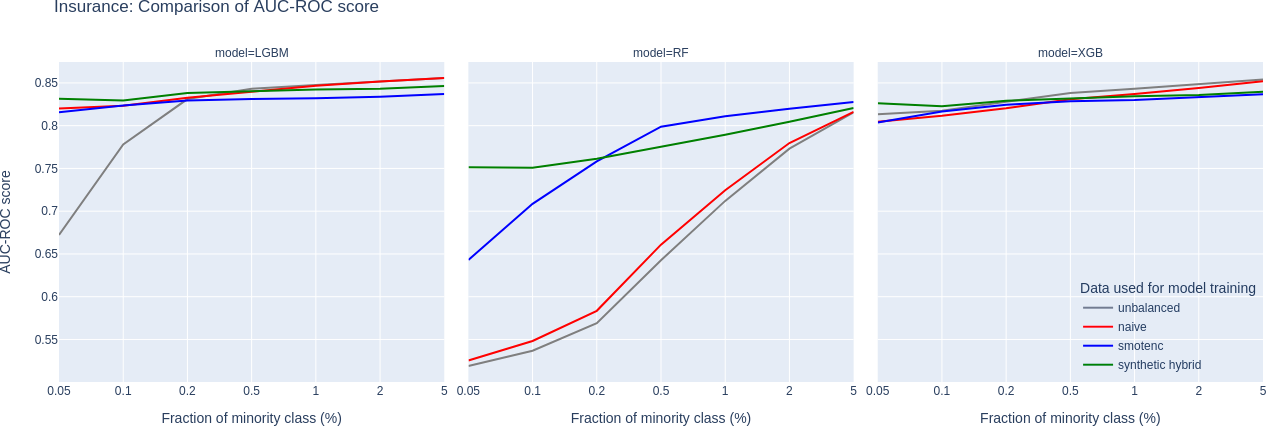}
        \caption{AUC-ROC}
    \end{subfigure}
    \hfill
    \begin{subfigure}{0.49\textwidth}
        \includegraphics[width=\linewidth]{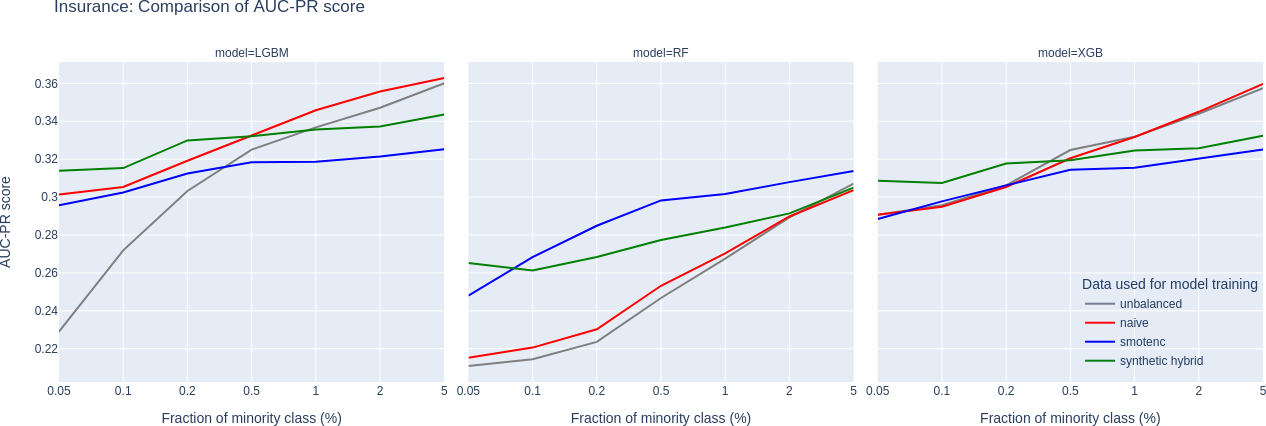}
        \caption{AUC-PR}
    \end{subfigure}
    \caption{AUC-ROC (a) and AUC-PR (b) of classifiers LGBM, RandomForest (RF), and XGB trained on the \textit{Insurance} data set to predict the target feature \textit{Response}. The classifiers are trained on unbalanced data sets (grey) and data sets that are upsampled naively (red), with the SMOTE-NC algorithm (blue), and with TabularARGN synthetic records (green). AUC values are reported for different fractions of the minority class in the unbalanced training data (x-axis).}
    \label{fig:insurance-auc-roc-pr}
\end{figure}

The Insurance data set is larger than \textit{Adult} and \textit{Credit Card}, resulting in a greater number of minority records (268) when downsampling to the 0.1\% minority fraction. This leads to a much more balanced performance across different upsampling techniques (Fig.~\ref{fig:insurance-auc-roc-pr}).

A notable difference in performance appears only for very small minority fractions. For minority fractions below 0.5\%, both the AUC-ROC and AUC-PR of LGBM and XGB classifiers trained on the TabularARGN hybrid data set are consistently higher than those for classifiers trained on other balanced data sets. However, the maximum performance gains are smaller than those observed for \textit{Adult} and \textit{Credit Card}.

\section{Conclusions}
Our benchmark study confirms the growing body of evidence that AI-based synthetic data generation is a practical and effective tool for addressing severe class imbalance in tabular data sets. By generating diverse, high-quality synthetic samples, this approach leads to improved predictive performance, particularly when the minority class is extremely underrepresented. Compared to traditional upsampling techniques like SMOTE-NC or naive oversampling, synthetic data upsampling consistently produces top-performing models across a variety of data sets and model types. Importantly, we demonstrate that such methods are now readily accessible through open-source libraries, such as the Synthetic Data SDK by MOSTLY AI, making them viable for real-world use with minimal configuration and computational effort. These findings support the continued adoption of synthetic upsampling as a robust, user-friendly solution for practitioners dealing with unbalanced data.

\bibliographystyle{plain}
\bibliography{references}

\end{document}